\documentclass[conference,letterpaper]{IEEEtran}
\usepackage{fancyhdr}
\usepackage[usenames,dvipsnames]{xcolor}

\usepackage{amsmath}
\usepackage[algoruled,vlined]{algorithm2e}
\usepackage{pgfplots}
\usepackage{subcaption}

\usepackage [english]{babel}
\usepackage [autostyle, english = american]{csquotes}
\MakeOuterQuote{"}

\pgfmathdeclarefunction{gauss}{2}{%
  \pgfmathparse{1/(#2*sqrt(2*pi))*exp(-((x-#1)^2)/(2*#2^2))}%
}

\begin{document}
\title{Towards Predicting First Daily Departure Times: \\ a Gaussian Modeling Approach for\\ Load Shift Forecasting}

\author{\IEEEauthorblockN{Nicholas H. Kirk}
\IEEEauthorblockA{Institute of Automatic Control Engineering\\ Technical University of Munich, Germany\\
        {\tt nicholas.kirk@tum.de}}
\and
\IEEEauthorblockN{Ilya Dianov}
\IEEEauthorblockA{Department of Informatics\\ Technical University of Munich, Germany}
{\tt ilya.dianov@tum.de}
}

\maketitle
\thispagestyle{plain}

\fancypagestyle{plain}{
\fancyhf{}	
\fancyfoot[L]{}
\fancyfoot[C]{}
\fancyfoot[R]{}
\renewcommand{\headrulewidth}{0pt}
\renewcommand{\footrulewidth}{0pt}
}

\pagestyle{fancy}{
\fancyhf{}
\fancyfoot[R]{}}
\renewcommand{\headrulewidth}{0pt}
\renewcommand{\footrulewidth}{0pt}

\begin{abstract}

This work provides two statistical Gaussian forecasting methods for predicting First Daily Departure Times (FDDTs) of everyday use electric vehicles. 
This is important in smart grid applications to understand disconnection times 
of such mobile storage units, for instance to forecast storage of
non dispatchable loads (e.g. wind and solar power).   
We provide a review of the relevant state-of-the-art driving behavior features towards FDDT prediction, to then propose an approximated Gaussian method which qualitatively forecasts how many vehicles will depart within a given time frame, by assuming that departure times follow a normal distribution. This method considers sampling sessions as Poisson distributions which are superimposed to obtain a single approximated Gaussian model.
Given the Gaussian distribution assumption of the departure times, we also model the problem with Gaussian Mixture Models (GMM), in which the priorly set number of clusters represents the desired time granularity. Evaluation has proven that for the dataset tested, low error and high confidence ($\approx 95\%$) is possible for 15 and 10 minute intervals, and that GMM outperforms traditional modeling but is less generalizable across datasets, as it is a closer fit to the sampling data.
Conclusively we discuss future possibilities and practical applications of the discussed model. 
\end{abstract}


\IEEEpeerreviewmaketitle

\section{Introduction}

With an increasing use of Plug-in Electric Vehicles (PEVs), mobile units can be seen as a potential grid-connected energy storage means without compromising their primary mobility functionality:
A PEV fleet can store, for instance, power from non dispatchable loads (e.g. solar panel and wind turbine sources) \cite{kempton2005vehicle}. However, connections of PEVs to the grid, in terms of times and locations, are complex to model given such logistic mobility. This work focuses on \textit{how to meaningfully model fleet-level departure times} over the commuter time frame 6 am - 9 am, in order to predict the availability of PEVs as grid storage over time. 
Heuristic assumptions such as over- or under-estimation of arrival/departure times both suffer from shortcomings and will result in inefficient energy use: an accurate forecast is therefore of paramount importance. We exploit First Daily Departure Times (FDDT), which are a key piece of information for connection time estimation in PEV load shifting algorithms \cite{goebel2013using},
but are hard to predict using historical realizations alone or via basic distribution modeling \cite{goebel2012forecasting}.
This research focused on understanding how to accurately predict PEV FDDT for successful load shift scheduling.
Such accuracy analysis was first performed via preliminary feature correlation analysis with FDDT (Section \ref{ssec:feature}), thanks to the availability of a dataset with diverse driving behavior features, which contain, for instance, information sampled from individual drivers regarding average trip length and duration (Section \ref{ssec:data_adopt}). Given the lack of forecasting capability of such features towards FDDT prediction, the research then makes progress towards approximated Gaussian modeling under specific a priori running assumptions (Section \ref{ssec:assumptions}). We provide theoretical background (Section \ref{sec:theoretical_framework}) to a method for computing lower and upper bounds of PEV FDDT for each time interval (Section \ref{ssec:comptimeint}), a time interval scaling method (Section \ref{ssec:granularity}), and provide a brief validation of such methods (Section \ref{sec:validation}).
Concludingly, we provide a summary on the proposed method and possible insights regarding future work (Section \ref{sec:conclusion}).

\section{Related Work}
Previous studies take into account aggregations of driver's behavior features for activity-based forecasting, aiming at Transportation Demand Management (TDM) \cite{kitamura1996applications}, congestion planning or logistic network optimality \cite{emmerink1995potential}.
In particular, behavior aggregation has been useful to understand the actions that provoke inter-relations among individuals, in order to cluster vehicle movement by activity \cite{kitamura1996applications}.

In \cite{ettema2003modeling} the FDDT prediction is based on utility maximization of the vehicle trip and activity participation. The activities are defined as driver intentions such as "being home before work", while trip is characterized by departure and arrival times.  
Another method \cite{chikaraishi2009exploring} uses a multilevel approach which claims that FDDT is dependent on individual attributes such as gender, age, profession and macro-level attributes such as day of the week, location and household income. Each attribute is modeled using normal distributions and the prediction is based on log likelihood maximization. However both these approaches require private (usually unavailable) information about each driver (e.g. type of activities engaged in after work). 
Goedel \cite{goebel2012forecasting} provides different approach which takes into consideration the day of the week as feature and a vehicle-based analysis of commuters, in order to predict a departure confidence interval. In other work \cite{ashtari@2012PEV}, charging profile predictions are based on stochastic analysis of the conditional Probability Density Function (PDF) over FDDT, daily arrival times and daily traveled distances. However, both methods provide a one hour interval precision of FDDT which is not sufficient within the domain of load shift prediction.
Given a low correlation among FDDT and driver behavior features, the presented research focused on Gaussian modeling of FDDT data only. Furthermore, the advantage of considering only \textit{first} daily departure times is that the research can disregard the complexity entailed by modeling the multiple stop factor.

\section{Data Understanding}
\label{sec:data_under}
We now describe the reasoning behind the adoption of the training and test set (Section \ref{ssec:data_adopt}), the feature correlation analysis (Section \ref{ssec:feature}), and the set of assumptions that are required for this statistical modeling problem (Section \ref{ssec:assumptions}).

\subsection{Data Adoption}
\label{ssec:data_adopt}
This project makes use of datasets from NREL's \textit{Secure Transportation Data Project} \cite{gonder2012establishing}, in particular \textit{Texas Department of Transportation - Transportation Studies with GPS Travel Diaries}.

The main reasons for such adoption are:

\begin{itemize}
  \item the dataset comprises many real-time features of the trips (e.g.
interval times, speeds, accelerations, statistical measures - see Table \ref{fig:NRELset}) 
  \item the features present high precision and low granularity
  \item all data has been electronically tracked
  \item given the geographical location (Texas), we assume climate variability to be low and therefore not influencing departure times
\end{itemize}

A major downfall of the dataset is that it does not comprise labeling for the day of the week, and furthermore all samplings have been performed only on Tuesdays or Wednesdays.

\begin{table}[h]
  \small
\begin{center}
    \begin{tabular}{ | p{3.3cm} | p{4.4cm} |}
    \hline
    Feature & Description \\ \hline
    {\sffamily start\_tm} & The start time of the first recorded point for the vehicle\\
    {\sffamily distance\_total} & Total travelled distance in miles\\
    {\sffamily percent\_fifty\_five\_sixty} & Percent of total time spent at speeds between fifty five and sixty miles per hour\\
    {\sffamily driving\_speed\_standard\newline \_deviation} & Standard deviation of driving speed distribution\\
    
    \hline
    \end{tabular}
\end{center}
\caption{Listing and descriptions of examples of features present in the NREL Transportation dataset \cite{gonder2012establishing}.}
\label{fig:NRELset}
\end{table}

\subsection{Feature Analysis}
\label{ssec:feature}

Correlation among potential features and the class to predict (FDDT) is a necessary but not sufficient condition for pattern learning.
We analyzed the potential predictive ability of each feature by executing a \textit{Correlation-based Feature Subset Selection} \cite{Hall1998} and a correlation-based Principal Component Analysis (PCA)
\cite{jolliffe2002principal}, making use of an Independent and Identically Distributed (IID) assumption. Such feature filters yielded a very low correlation between features and data, making these unserviceable for machine learning (see Table \ref{fig:lowcorrresults}).

\begin{table}[h]
 \small
\begin{center}
    \begin{tabular}{ | l | p{1.5cm} |}
    \hline
    Correlation-based selected features & Correlation with FDDT ({\sffamily start\_tm}) \\ \hline
    {\sffamily total\_speed\_velocity\_ratio} & +0.15 \\
    {\sffamily percent\_distance\_fifty\_five\_sixty} & -0.21\\
    {\sffamily absolute\_time\_duration\_hrs} & -0.3\\
    {\sffamily descending\_rate\_median\_absolute\_deviation} & -0.08\\
    {\sffamily max\_deceleration\_event\_duration} & -0.33\\
    {\sffamily average\_acceleration\_event\_duration} & +0.04\\
    {\sffamily min\_deceleration\_event\_duration} & -0.05\\    
    \hline
    \end{tabular}
    
\end{center}
\caption{Features with the highest correlation with FDDT ({\sffamily start\_tm}), and with the lowest correlation among themselves. For a description of the cited features, see \cite{gonder2012establishing}.}
\label{fig:lowcorrresults}
\end{table}

\subsection{Assumptions}
\label{ssec:assumptions}
An initial intuition after viewing the variety of available features in the dataset (Section
\ref{ssec:data_adopt}) would suggest the possibility of performing high-dimensional regression with such diverse components. However, this approach is not possible with the current dataset, since features presented very low correlation values and hence low or no learning potential (see Table \ref{fig:lowcorrresults}).
Therefore we proceed in assuming that every sample is Independent and Identically Distributed (IID), i.e. that the FDDT of a vehicle does not influence the FDDT of another.
Consequently, we do not consider the problem as a \textit{time-series analysis} as understood in literature \cite{box2013time}.\\
Given such information, instead of predicting the exact departure time of the sample, it is more convenient to forecast, given historical values, i) how many FDDT will fall within certain time intervals, ii) the confidences of the latter, and iii) a system-level interval granularity itself. By empirical analysis and by assumption we define that the FDDT sampling undertaken for the training dataset is distributed according to Poisson's definition. 

\section{Approximated Gaussian Modeling}
\label{sec:approxgaussmod}
We proceed in describing a method to constrain our problem to the Gaussian modeling domain (Section \ref{ssec:comptimeint}, \ref{ssec:granularity}),
given the assumptions in Section \ref{ssec:assumptions}.

\subsection{Theoretical Framework}
\label{sec:theoretical_framework}
\subsubsection{Poisson distribution}
\label{poisson}
In probability theory, the discrete Poisson distribution expresses the
likelihood of a number of events occurring sequentially and independently of each other within a given
time frame, knowing that on average a given number $\lambda$ occurs. For an in-depth description of the mathematical properties of this distribution, we refer to \cite{haight1967handbook}.\\ 
We exploit the mathematical property that for a hypothetically infinite number of samplings, the superimposition of such Poisson distributions converges to a Gaussian (Normal) distribution \cite{haight1967handbook}. 
Due to the latter property, it is then possible to model with traditional Gaussian assumptions.
\subsubsection{Gaussian distribution}
\label{gaus_distr}
A Gaussian (Normal) is a continuous probability distribution that often characterizes real-valued random variables in applied contexts. In this context we model the distribution over time intervals and their confidence. The latter is possible by defining the percentage of values captured by a distance $k\sigma$, where $\sigma^2$ is variance from the mean $\mu$, as seen in (\ref{eq:gassiandensityfunc}). 

\begin{align}
\label{eq:gassiandensityfunc}
\int_{-k\sigma}^{+k\sigma} \mathcal{N}(\mu,\sigma^2) 
\end{align}

For $k = 2$ we obtain confidence of $\approx 95\%$.
For a deeper mathematical description we refer to \cite{dudley2002real}.

\subsection{Computing Time Intervals}
\label{ssec:comptimeint}

By aggregating the timestamp samples of a single sampling session in
discrete time intervals, we obtain a Poisson distribution.\\
If we sample a \textit{sufficiently large dataset}, or a heterogeneous set of sampling sessions, the superimposition of these Poisson distributions converge to an approximated Gaussian distribution \cite{haight1967handbook}.\\
Let {$n$} be the index of the current sampling session and $N$ the total number of samplings which have been operated. Let {$b$} be an arbitrary constant that defines the number of bins (and therefore the time interval granularity).
We then construct a matrix $K$:\\ 
\begin{equation}
  K^b_{n} =
 \begin{pmatrix}
  K^0_{0}   & \cdots & K^b_{0} \\
  K^0_{1} & \cdots & K^b_{1} \\
  \vdots  & \ddots & \vdots  \\
  K^0_{n} & \cdots  & K^b_{n}
 \end{pmatrix}
\end{equation}

We compute an estimation of the lower bound and the upper bound of the number of PEV departures in the time interval $t_i \in 0, \ldots, b$ with a probability of 95\%:
\begin{align}
\text{\small time interval margins }t_i = \left[m^i - 2\sqrt{m^i},~ m^i + 2\sqrt{m^i}\right]  
\end{align}
where: 
\begin{align}
  m^i = \frac{1}{n} \sum\limits_{j=0}^{n} K^i_j
\end{align}

\subsection{Granularity Scaling}
\label{ssec:granularity}
We want to hypothetically increase $b$ in order to have a time prediction interval as small as possible. 
We define the error $\varepsilon_{\%}$ as the wanted percentage error of our confidence interval.

We obtain the lowest time granularity possible without lowering the given confidence interval by imposing that: 
\begin{align}
\varepsilon_{\%} \leq \frac{m^{min}}{\sqrt{m^{min}}}
\end{align}
 
where $0 \leq \varepsilon_{\%} \leq 1$ and: 
\begin{align}
m^{min} = \min( m^0 \dots m^b)
\end{align} 
An overview of the entire modeling here discussed in Sections \ref{ssec:comptimeint} and \ref{ssec:granularity} can be viewed in Algorithm 1.

{
\begin{algorithm}[h]
\label{alg:PEVInIntCount}
\DontPrintSemicolon
 \KwData{
 \begin{itemize}
   \item $minDepTime$, earliest departure time
   \item $maxDepTime$, latest departure time
   \item $TD$, a training set containing $n$ sampling sessions of departure times  
   \item $CIvalue$, the percentage value of the desired confidence interval
 \end{itemize}
 }
\vspace{1mm}
 \KwResult{$TimeIntMargins$, PEV departure number for each time interval}
\vspace{1mm}
\Begin{
     $b$ $\longleftarrow$ $0$\;
     $j$ $\longleftarrow$ $0$\;
     $Mmatrix$ $\longleftarrow$ $0$\;
     $\epsilon_\%$ $\longleftarrow$ $\left(1 - CIvalue\right)$\;
     $TD_{cut}$ $\longleftarrow$ $trimRange(TD,minDepTime,maxDepTime)$\;
     \While{$\epsilon_\% < \left(min_i\left(Mmatrix^i\right)/\sqrt{min_i\left(Mmatrix^i\right)}\right)$}{
         \textbf{increase} $b$\;
         $Kmatrix$ $\longleftarrow$ $divideInIntervals(TD_{cut},b)$\;
         $Mmatrix$ $\longleftarrow$ $imposeAndAvg(Kmatrix,b,n)$\;
     }
     \While{$j < b$}{
             $intMargins^j$ $\longleftarrow$ $compMargins(Mmatrix^j)$\;
             \textbf{output} $intMargins^j$\;
             \textbf{increase} $j$\;
     }
}
\caption{PEV departure number within time interval computation}
\end{algorithm}
}

\subsection{Expectation-Maximization for Gaussian Mixture Models}
\label{em_gmm}
Given the Gaussian assumption used throughout this text, for which all first time departures follow a Normal distribution, we made use of Gaussian Mixture Models (GMM) to cluster FDDTs, in which each cluster represents a bin as described in Section \ref{ssec:comptimeint}.
We model time intervals as a Gaussian distribution, and the time inside each interval is additionally characterized by a Gaussian distribution. For this we use a mixture model with K components where each component is a multivariate Gaussian density:
\begin{equation}
g_i(x| \mu_i, \Sigma_i)= \frac{1}{ (2\pi)^{D/2}|\Sigma_i|^{1/2}}e^{- \frac{1}{ 2}(x-\mu_i)^T\Sigma_i^{-1}(x-\mu_i)}
\end{equation}
where $\mu_i$ is a mean, $\Sigma_i$ is a covariance matrix, $x \in D$, where $D$ is a given dataset, $i=1, \ldots, K$.
In order to learn unsupervisedly the parameters of the latent models characterizing the multivariate Gaussian distribution, we make use of the iterative Expectation Maximization (EM) algorithm \cite{dempster1977maximum}, which finds the maximum likelihood of parameters also with low resolution distribution data, such as in the case of our approximated Gaussian model derived from superimposed Poisson distributions.
For a more detailed description of EM for GMM, we refer to \cite{dempster1977maximum}.

\section{Validation Results and Model Usability}
\label{sec:validation}

\begin{figure}[h]
\centering
\scalebox{1.3}{
\includegraphics{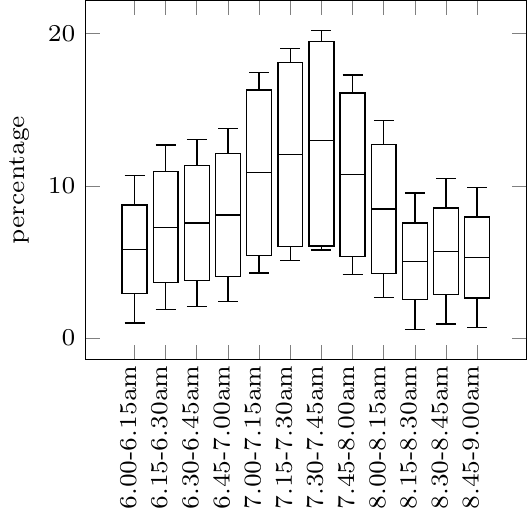}
}
\caption{Graphical representation of the confidence intervals obtained with the approximated Gaussian modeling described in Section \ref{ssec:comptimeint}.}
\label{fig:margin_boxplot}
\end{figure}

\begin{figure}[t]
\centering
\includegraphics[width=0.5\textwidth]{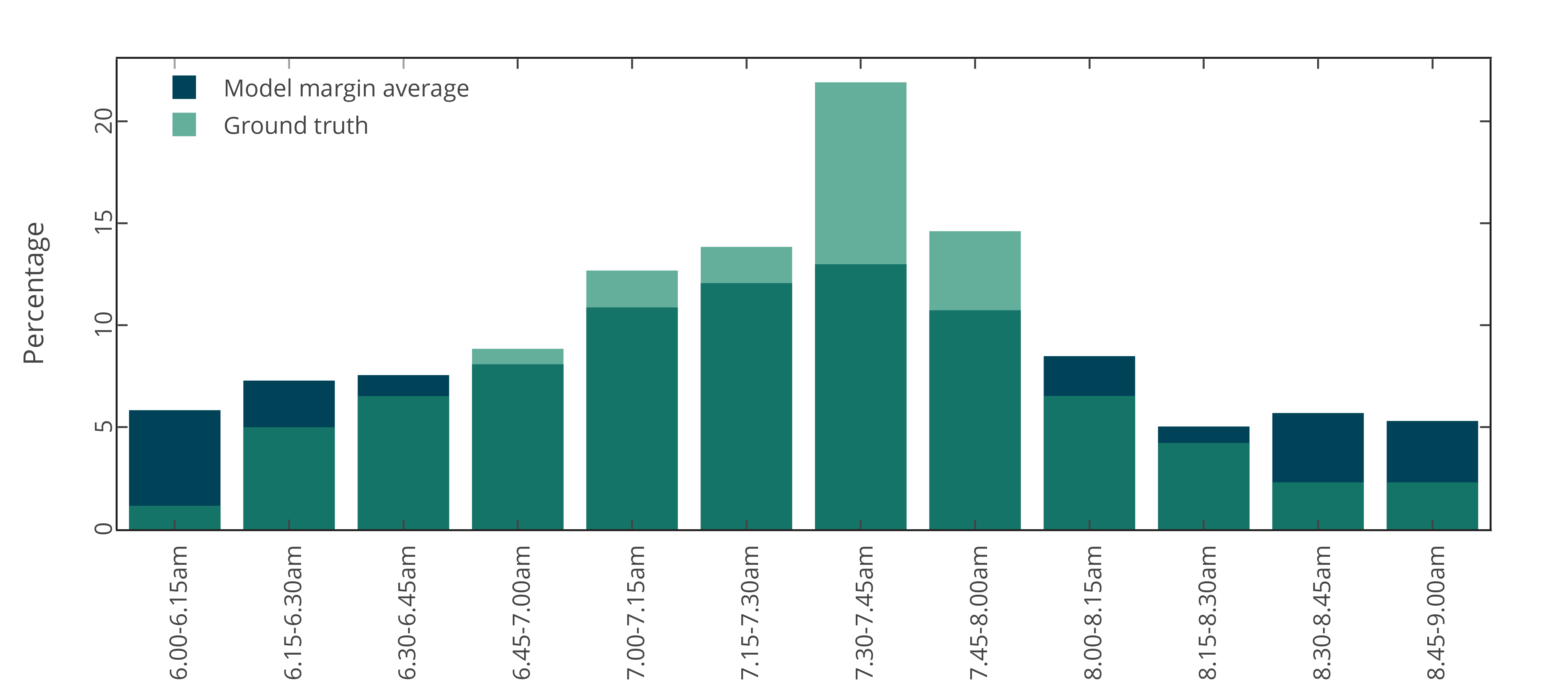}
\caption{The superimposed Gaussian model validated against \textit{Rio Grande Valley} ground values (12 bins, 15 minutes each).}  
\label{fig:validation2}
\end{figure}

Validation was not possible with the current running assumptions, given the reduced number of samples of our training data after time frame ($6am-9am$) pruning.
In order to increase the number of training samples, we superimposed the pruned sets of different cities which share similar urban and climatic characteristics (namely \textit{Austin, San Antonio, Houston, El Paso}) to create a training set, and then used the superimposed Gaussian model to validate on a test set formed by samples from
\textit{Rio Grande Valley}. The training set for \textit{all three incoming experiments} contained 758 instances of data for the time period from 6:00 am to 9:00 am, and the test set for all experiments contained 260 instances of data for the same time period. The set was normalized for explanatory convenience (i.e. every bin defines the percentage of vehicles that depart in the bin time interval). 
Statistical dispersion was computed with Gauss error function (erf) for our validating model which evaluates the probability that measurement $x$ is within a range from 
$-\frac{x}{\sigma\sqrt{2}}$ to $\frac{x}{\sigma\sqrt{2}}$. To compute erf we used a following formula:
\begin{align}
erf(x)=\frac{x}{\sqrt{\pi}}\int\limits_0^x e^{-t^2}\mathrm{d}t
\end{align}
\paragraph{First experiment (approximated Gaussian modeling, 12 timeframes)} The model derived from the theory presented in Section \ref{ssec:comptimeint} validated the model on all bins. Margin computations are shown in Fig. \ref{fig:margin_boxplot}, while validation results can be seen in Fig. \ref{fig:validation2}. 
The resulting erf for such approximated Gaussian modeling is shown in Table \ref{table_erf_quasi}.

\begin{table}[t]
\scriptsize
\begin{center}
\begin{tabular}{|c|c|c|}
\hline
 Time intervals & Average margin values & erf values\\
\hline
6.00-6.15am & 0.0584 & 0.0658 \\
\hline
6.15-6.30am & 0.0729 & 0.0821 \\
\hline
6.30-6.45am & 0.0756 & 0.0851 \\
\hline
6.45-7.00am & 0.0809 & 0.0911 \\
\hline
7.00-7.15am & 0.1088 & 0.1223 \\
\hline
7.15-7.30am & 0.1207 & 0.1355 \\
\hline
7.30-7.45am & 0.1300 & 0.1459 \\
\hline
7.45-8.00am & 0.1074 & 0.1207 \\
\hline
8.00-8.15am & 0.0849 & 0.0956 \\
\hline
8.15-8.30am & 0.0504 & 0.0568 \\
\hline
8.30-8.45am & 0.0570 & 0.0642\\
\hline
8.45-9.00am & 0.0531 & 0.0599 \\
\hline
\multicolumn{2}{|c|}{Average erf value} & 0.0938 \\
\hline
\multicolumn{2}{|c|}{Normalized score on a number of bins} & 1.1256 \\
\hline
\end{tabular}
\end{center}
\caption{Gauss error function results for the approximated Gaussian model on average margin values (12 bins, 15 minutes each).}
\label{table_erf_quasi}
\end{table}

\begin{table}[t]
\scriptsize
\begin{center}
\begin{tabular}{|c|c|c|}
\hline
  Time intervals & Predicted values & erf values\\
\hline
6.00-6.15am & 0.0269 & 0.0303 \\
\hline
6.15-6.30am & 0.0654 & 0.0737 \\
\hline
6.30-6.45am & 0.1038 & 0.1167 \\
\hline
6.45-7.00am & 0.0692 & 0.0780 \\
\hline
7.00-7.15am & 0.1577 & 0.1765 \\
\hline
7.15-7.30am & 0.0962 & 0.1082 \\
\hline
7.30-7.45am & 0.0146 & 0.0165 \\
\hline
7.45-8.00am & 0.0135 & 0.0152 \\
\hline
8.00-8.15am & 0.0692 & 0.0780 \\
\hline
8.15-8.30am & 0.0769 & 0.0866 \\
\hline
8.30-8.45am & 0.0346 & 0.0390 \\
\hline
8.45-9.00am & 0.0192 & 0.0217 \\
\hline
\multicolumn{2}{|c|}{Average erf value} & 0.0700 \\
\hline
\multicolumn{2}{|c|}{Normalized score on a number of bins} & 0.8400 \\
\hline
\end{tabular}
\end{center}
\caption{Gauss error function results for GMM-model on \textit{Rio Grande Valley} values (12 bins, 15 minutes each).}
\label{table_erf_em}
\end{table}

\begin{figure}[t]
\centering

\includegraphics[width=0.5\textwidth]{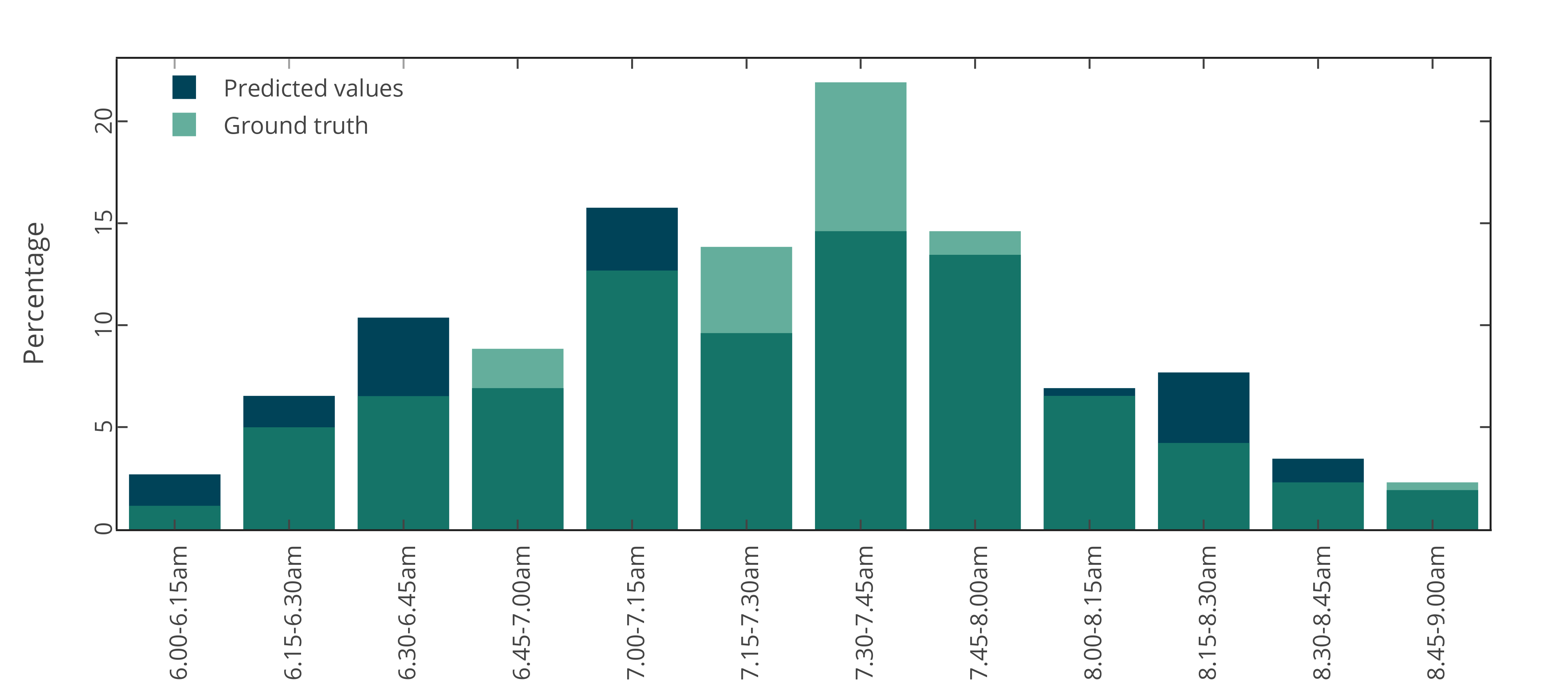}
\caption{EM for Gaussian Mixture Model algorithm tested on \textit{Rio Grande Valley} dataset (12 bins, 15 minutes each).}  
\label{fig:validationEMjustTime}
\end{figure}

The implementation of Algorithm 1 focused on granularity understanding and margin computation, in which we can compute a trade-off between estimation confidence (via modeling the $k$ parameter in Eq. \ref{eq:gassiandensityfunc}), and time interval granularity (i.e. the number of bins).
\paragraph{Second experiment (Gaussian Mixture Modeling, 12 timeframes)} For the EM method we used training data containing only FDDTs ({\sffamily start\_tm}) for different vehicles as feature and validated it on the previously mentioned test data (\textit{Rio Grande Valley} set). Each instance of the dataset is associated with a single vehicle and the resulting model of the EM algorithm, shown in Fig. \ref{fig:validationEMjustTime}, illustrates the dependency between departure times and number of vehicles departing at the particular time interval. We modeled 12 clusters, i.e. 12 timeframes of 15 minutes each.
The model shows that the highest amount of vehicles was departing at 7:00 am - 7:15 am which makes this interval the most probable for future predictions under the current assumptions. The resulting erf values are shown in Table \ref{table_erf_em}. The average erf values are 0.094 and 0.07 for approximated-Gaussian modeling and EM for GMM respectively (i.e. first and second experiment).

\begin{figure}[t]
\centering
\includegraphics[width=0.5\textwidth]{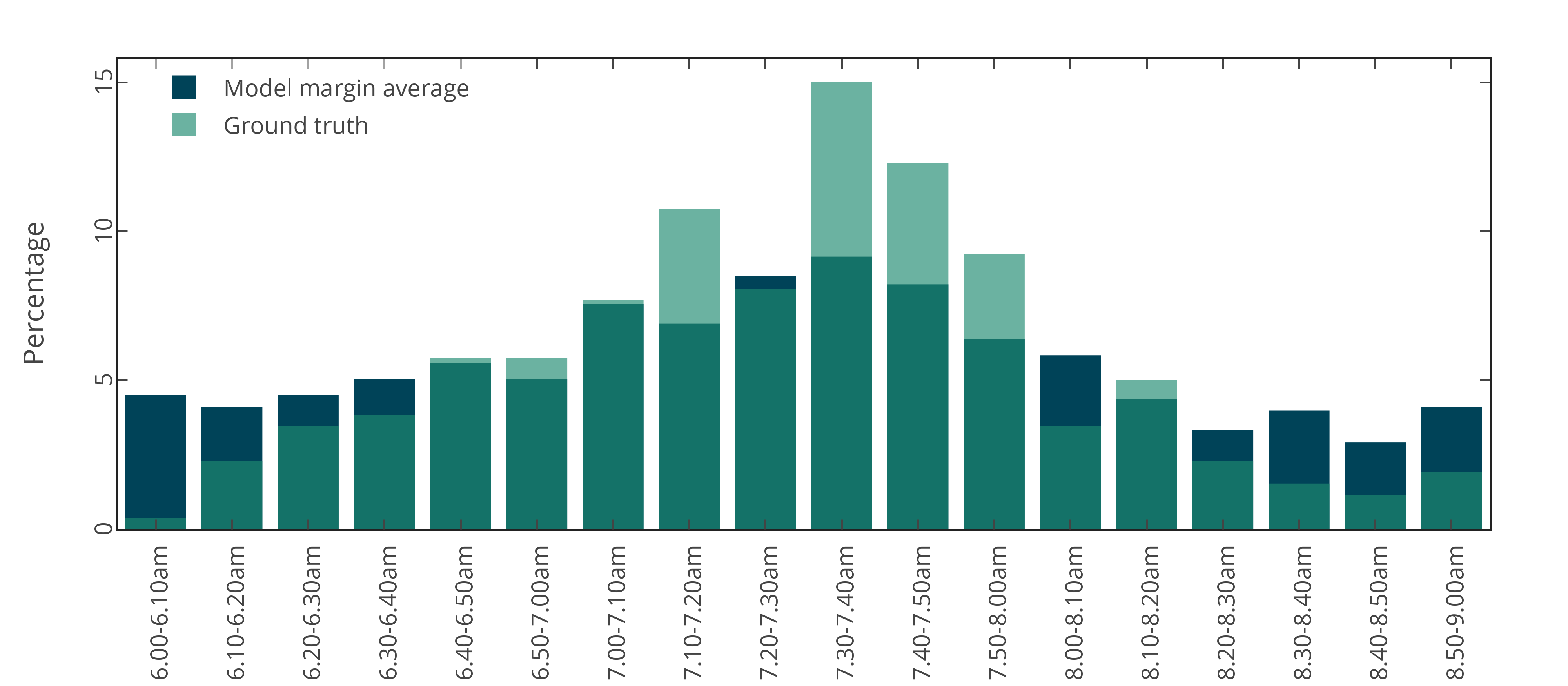}
\caption{The superimposed Gaussian model validated against \textit{Rio Grande Valley} ground values (18 bins, 10 minutes each).}
\label{fig:approxgauss_for_10}
\end{figure}

\begin{figure}[t]
\centering
\includegraphics[width=0.5\textwidth]{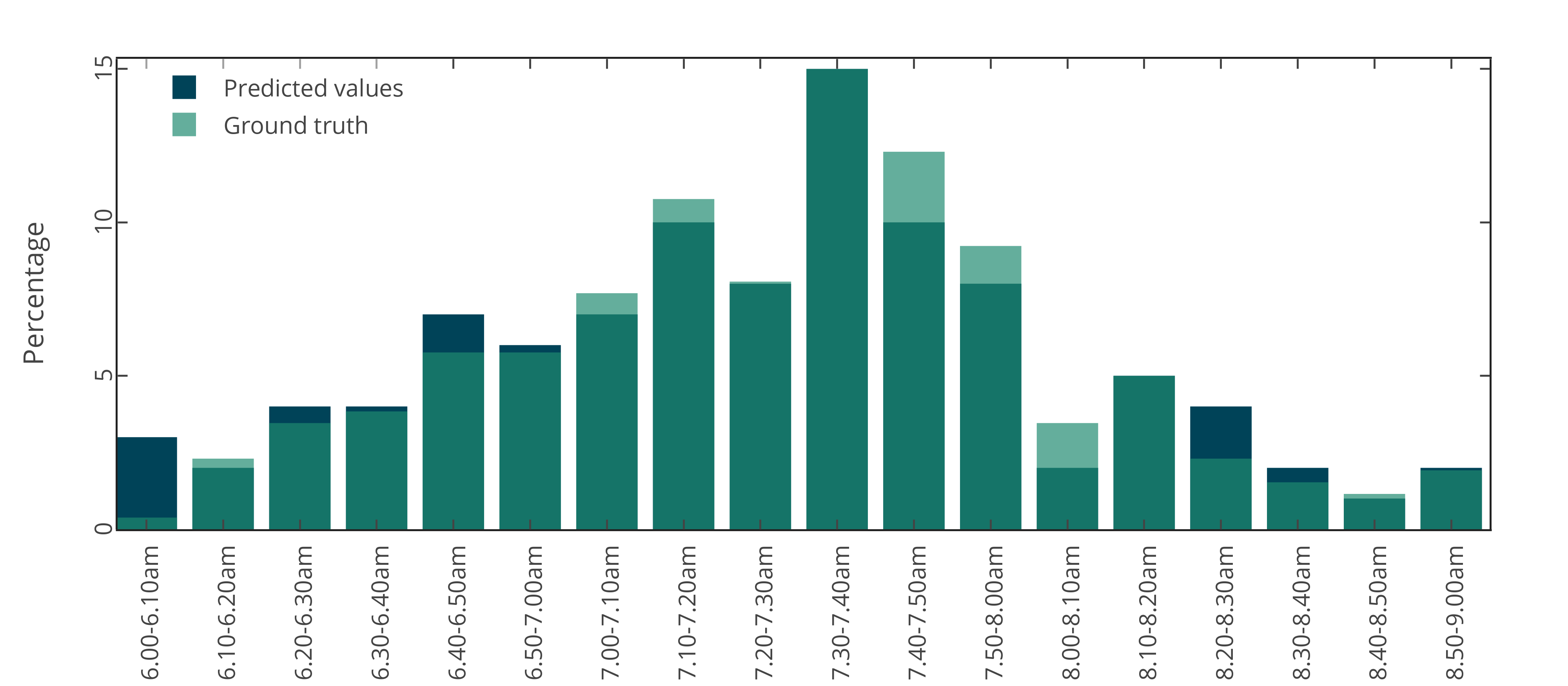}
\caption{EM for Gaussian Mixture Model algorithm tested on \textit{Rio Grande Valley} dataset (18 bins, 10 minutes each).}
\label{fig:em_for_10}
\end{figure}

\paragraph{Third experiment (Traditional and Gaussian Mixture Modeling, 18 timeframes)} We repeated both the EM method for Gaussian Mixture Modeling and traditional approximated Gaussian modeling, this time with 18 timeframes, i.e. 10 minute intervals, using only FDDTs ({\sffamily start\_tm}). Results are visible in Fig. \ref{fig:approxgauss_for_10} and \ref{fig:em_for_10}, while Gaussian error values are displayed in Table \ref{table_erf_em_10} and \ref{table_erf_ag_10}. 
Overall, given the graphical and error results, we can confirm the Gaussian assumption on such real data model. We can observe that approximated Gaussian modeling preserves the form factor across departures per time intervals, whereas GMM provide a closer approximation to the training data.

\begin{figure*}[t]
\centering
\includegraphics[width=\textwidth]{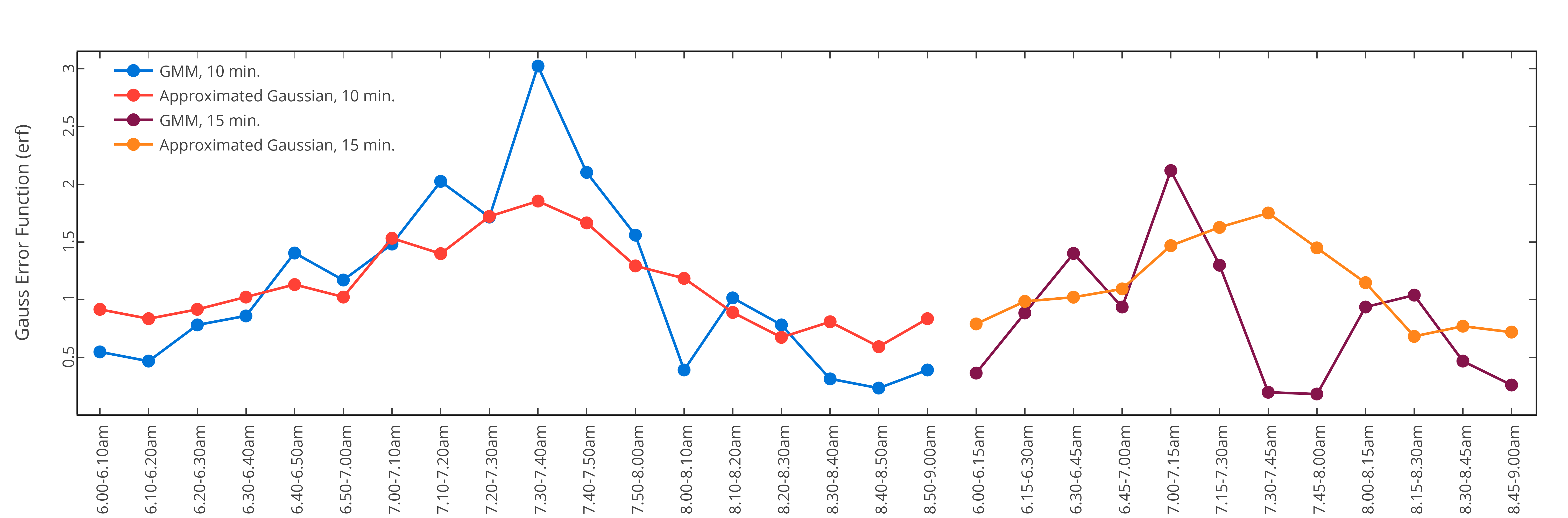}
\caption{Gauss error function (erf), normalized over the number of bins, across both GMM and approximated Gaussian models for 10 minute time intervals (left), and 15 minute intervals (right).}
\label{fig:fullerror}
\end{figure*}

\section{Conclusions and Future Work}
\label{sec:conclusion}

Modeling First Daily Departure Times (FDDT) of electric vehicles is of paramount importance for smart grid load shift planning, as these can be used as temporary energy storage units. By making a Gaussian distribution assumption of such departure times, we have provided a i) traditional Gaussian modeling approach with confidence and time interval size modeling, and ii) a Gaussian Mixture Model approach to compute clusters associated to time intervals.  
Evaluation has proven that for the dataset tested, low error and high confidence ($\approx 95\%$) is possible for 15 and 10 minute intervals. By inspection of the normalized score on a number of bins for both methods (Fig. \ref{fig:fullerror}), we notice that GMM method is more subject to error when increasing time interval granularity, but requires less data to formulate the model.
Future work will be oriented towards testing the presented 
Gaussian model on large datasets, implementing error propagation when 
relaxing the IID assumption (i.e. assuming that all cars depart), and considering confidence and error trade-off for practical applications.  
Furthermore a collaboration with transport survey research centers
would be useful to gather more vehicle-related and activity-related data, in order to cluster by the latter and by points of interest, to then verify feature correlation with FDDTs.

\section*{Acknowledgements}
The authors thank Christoph G\"{o}bel (Technical University of Munich, Germany) for his valuable feedback on an initial version of this paper.

\begin{table*}[t]
\scriptsize

\parbox{.45\linewidth}{
\begin{center}
\begin{tabular}{|c|c|c|}
\hline
  Time intervals & Predicted values & erf values\\
\hline
6.00-6.10am	&	0.0269	&	0.0304 \\
\hline
6.10-6.20am	&	0.0231	&	0.0260\\
\hline
6.20-6.30am	&	0.0385	&	0.0434\\
\hline
6.30-6.40am	&	0.0423	&	0.0477\\
\hline
6.40-6.50am	&	0.0692	&	0.0780\\
\hline
6.50-7.00am	&	0.0577	&	0.0650\\
\hline
7.00-7.10am	&	0.0731	&	0.0823\\
\hline
7.10-7.20am	&	0.1000	&	0.1125\\
\hline
7.20-7.30am	&	0.0846	&	0.0953\\
\hline
7.30-7.40am	&	0.1500	&	0.1680\\
\hline
7.40-7.50am	&	0.1038	&	0.1168\\
\hline
7.50-8.00am	&	0.0769	&	0.0866\\
\hline
8.00-8.10am	&	0.0192	&	0.0217\\
\hline
8.10-8.20am	&	0.0500	&	0.0564\\
\hline
8.20-8.30am	&	0.0385	&	0.0434\\
\hline
8.30-8.40am	&	0.0154	&	0.0174\\
\hline
8.40-8.50am	&	0.0115	&	0.0130\\
\hline
8.50-9.00am	&	0.0192	&	0.0217\\
\hline
\multicolumn{2}{|c|}{Average erf value} & 0.0625 \\
\hline
\multicolumn{2}{|c|}{Normalized score on a number of bins} & 1.1250 \\
\hline
\end{tabular}
\end{center}
\caption{Gauss error function results for the GMM model validated against \textit{Rio Grande Valley} values (18 bins, 10 minutes each).}
\label{table_erf_em_10}
}
\hfill
\parbox{.45\linewidth}{
\scriptsize
\begin{center}
\begin{tabular}{|c|c|c|}
\hline
  Time intervals & Predicted values & erf values\\
  \hline
6.00-6.10am	&	0.0451	&	0.0509 \\
\hline				
6.10-6.20am	&	0.0411	&	0.0464 \\
\hline				
6.20-6.30am	&	0.0451	&	0.0509 \\
\hline				
6.30-6.40am	&	0.0504	&	0.0568 \\
\hline				
6.40-6.50am	&	0.0557	&	0.0628 \\
\hline				
6.50-7.00am	&	0.0504	&	0.0568 \\
\hline				
7.00-7.10am	&	0.0756	&	0.0851 \\
\hline				
7.10-7.20am	&	0.069	&	0.0777 \\
\hline				
7.20-7.30am	&	0.0849	&	0.0956 \\
\hline				
7.30-7.40am	&	0.0915	&	0.1030 \\
\hline				
7.40-7.50am	&	0.0822	&	0.0925 \\
\hline				
7.50-8.00am	&	0.0637	&	0.0718 \\
\hline				
8.00-8.10am	&	0.0584	&	0.0658 \\
\hline				
8.10-8.20am	&	0.0438	&	0.0494 \\
\hline				
8.20-8.30am	&	0.0332	&	0.0374 \\
\hline				
8.30-8.40am	&	0.0398	&	0.0449 \\
\hline				
8.40-8.50am	&	0.0292	&	0.0329 \\
\hline				
8.50-9.00am	&	0.0411	&	0.0464 \\
\hline
\multicolumn{2}{|c|}{Average Gaussian error (erf) value} & 0.0626 \\
\hline
\multicolumn{2}{|c|}{Normalized score on a number of bins} & 1.1268 \\
\hline
\end{tabular}
\end{center}

\caption{Gauss error function results for Approximate Gaussian model validated against \textit{Rio Grande Valley} values (18 bins, 10 minutes each).}
\label{table_erf_ag_10}
}
\end{table*}

\IEEEtriggeratref{14}

\bibliography{main}{}
\bibliographystyle{IEEEtran}

\end{document}